\documentclass[10pt,twocolumn,letterpaper]{article}
\usepackage[pagenumbers]{style}
\usepackage{cuted}
\usepackage{caption}
\usepackage{booktabs}
\usepackage{multirow}
\usepackage{comment}
\usepackage{amsmath,amssymb}
\usepackage{graphicx}
\usepackage{xspace}
\definecolor{linkblue}{rgb}{0.21,0.49,0.74}
\usepackage[pagebackref,breaklinks,colorlinks,allcolors=linkblue]{hyperref}
\usepackage[capitalize,nameinlink]{cleveref}
\usepackage{enumitem}
\setlist[itemize]{topsep=3pt,itemsep=1pt,parsep=0pt,partopsep=0pt}

\crefname{figure}{Figure}{Figures}
\Crefname{figure}{Figure}{Figures}
\crefname{table}{Table}{Tables}
\Crefname{table}{Table}{Tables}
\crefname{section}{Section}{Sections}
\Crefname{section}{Section}{Sections}
\AtBeginDocument{\crefname{figure}{Figure}{Figures}\crefname{table}{Table}{Tables}\crefname{section}{Section}{Sections}}

\newcommand{\papertitle}{Budgeted Safety Verification for Streaming Video Generation}
\title{\papertitle}

\author{Yuxin Cao\textsuperscript{1} \quad Ziqi Ding\textsuperscript{2} \quad Jiayu Zhang\textsuperscript{3} \quad Xianglin Yang\textsuperscript{1} \quad Wei Song\textsuperscript{4} \quad Jin Song Dong\textsuperscript{1}\\[0.5em]
\textsuperscript{1}National University of Singapore, Singapore\\
\textsuperscript{2}University of New South Wales, Australia\\
\textsuperscript{3}The University of Hong Kong, Hong Kong SAR, China\\
\textsuperscript{4}Griffith University, Australia}

\begin{document}
\maketitle

\begin{abstract}
A streamed video generator releases each chunk the moment that it is produced, so a safety decision that waits for the finished clip arrives after the harm has been seen. We measure a guardrail by exposure, the harmful frames a viewer is shown, and ask where a fixed moderation budget should be spent. The accurate detector reads decoded frames with a video language model and is too expensive for every chunk, so our guard reads hidden states the generator has computed already and spends a cheap score on deciding when to call it. On 599 streams from a distilled causal generator the undefended model shows 46.25 harmful frames of 81. The video model on every chunk leaves 0.91 frames at 80.1 scored per benign stream, and the same model triggered by our guard leaves 2.35 at 5.3. Spending that budget at random leaves 37.20 frames, so the gain is in the ordering, and a frame detector in the same role gives up almost six, so the judging stage must read the chunk as a clip. The ordering is worth more as the stream lengthens, and the guard carries to a second system. An independent guardrail model agrees with our oracle at $\kappa=0.74$, and a blinded audit by 50 participants puts its precision at 96.1\% and recall at 82.8\%.
\textcolor{red}{\textbf{Warning.} This paper contains examples of generated content that some readers may find disturbing or offensive in nature.}
\end{abstract}

\begin{figure*}[!t]
  \centering
  \includegraphics[width=\textwidth]{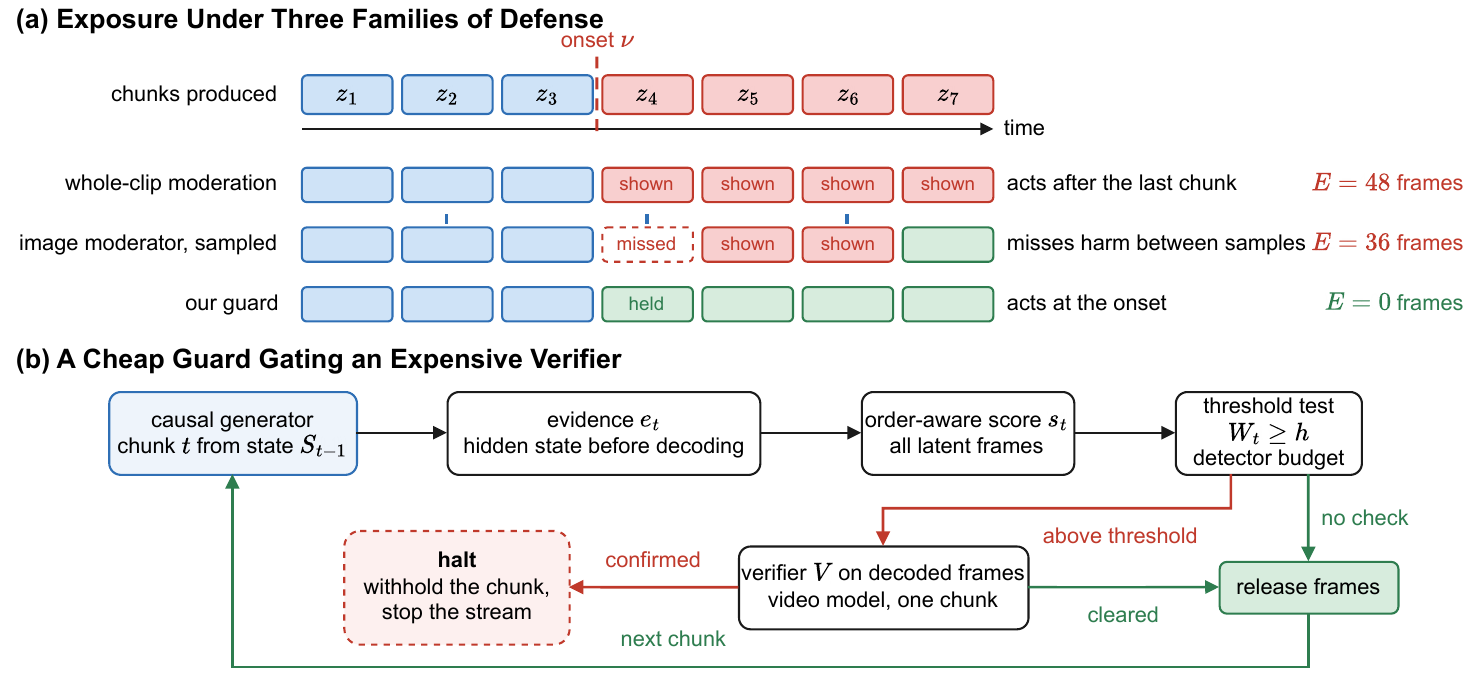}
  \caption{Budgeted safety verification of a streamed video generator. Only a chunk that the guard flags is shown to the verifier.}
  \label{fig:overview}
\end{figure*}

\section{Introduction}
\label{sec:intro}
Video generation has become a streaming service. A distilled causal model produces a clip chunk by chunk and releases each chunk as soon as the decoder turns it into frames, which makes a generator behave like a live feed and lets a user steer the scene while it is still being produced~\cite{huang2025selfforcing,yin2025slow,yang2026longlive,feng2025streamdiffusionv2}. Interactive products are built on exactly this property, since a viewer who waits for a finished clip cannot redirect the scene and a generator that answers in chunks is the only kind that lets them. Safety is still organized around a finished file. A text filter screens the prompt, the model runs to completion, and a moderator inspects the clip before anyone publishes it, so every stage assumes that no frame reaches a viewer until the last frame exists. Streaming removes that assumption and changes what a defense has to accomplish.

Two properties separate this setting from image safety. Firstly, release is irreversible. An image moderator that fires after the pixels are drawn still blocks them before anyone sees them, while a streamed chunk is public the moment that it is decoded, so a defense is measured by how many harmful frames it allowed through before acting, and catching the harm afterwards settles nothing. Secondly, harm can live in the ordering of the frames rather than in any one of them. A hand reaching toward a stove, a figure stepping over a railing and a crowd closing around a person are each unremarkable as a still, and what makes the sequence unsafe is the order that the frames arrive in. Filters derived from image models miss exactly this content, because a score that pools frames symmetrically cannot separate a clip from a shuffle of that clip~\cite{cao2026temporal}. Sampling widens the gap, since the blind interval between two inspected frames grows as the sampling rate falls~\cite{cao2026failures}. Existing defenses inherit the finished-file assumption. Prompt-level filtering reads the request alone, so it is blind to a request that does not state its harm and does not run at all in the open-weight deployments the generators studied here are distributed under. Frame-level moderation concedes at least the frames it is still examining and is blind to order by construction, whole-clip moderation reaches its verdict too late for a live feed~\cite{chen2025safewatch}, and generation-time defenses rewrite a single joint denoising pass on the presumption that none of the clip has been shown~\cite{yoon2025safree,kundu2026reins}. None of them reports what a viewer has already seen, and none of them can, because that quantity is undefined for any method whose output does not come into existence until the whole of the clip is finished.

We treat the problem as a decision, taken chunk by chunk, about whether to release the chunk and whether to pay for a closer look at it. The generator exposes evidence about a chunk before the decoder runs, so a guard reading a mid-depth hidden state can act at the onset at no cost in exposure, and \cref{fig:overview} shows the loop and what each family of defense concedes on one stream. We score each chunk with a small temporal head over its latent frames. That score is cheap enough for every chunk and too weak to be trusted alone, which is the situation a cascade is built for, so we spend it on deciding when an expensive decoded-pixel verifier is worth a call. The division of labor decides where sensitivity to order has to live, and our measurements put it in the expensive stage, since a frame detector in place of the video model gives up almost six frames of exposure at the same budget. The threshold itself is fixed by conformal risk control on benign streams, so it carries a distribution-free bound on how often an exchangeable benign stream ever crosses it, where a crossing triggers a verifier call and leaves the stream running, and an operator sets it by naming the detector cost they are willing to pay. That threshold then sets a compute budget rather than an interruption rate, because a chunk the verifier clears is released and the stream continues, and that asymmetry lets the threshold fall far below anything a halting rule on its own could tolerate. The composition releases 2.35 harmful frames of the 46.25 an undefended generator releases, while scoring a fifteenth of the benign frames that the video model on every chunk has to score. The same budget spent on randomly chosen chunks leaves 37.20 frames, so almost all of the gain lies in the ordering. The value of the ordering grows with the stream, since tripling its length roughly triples what the guard is worth to a deployment, and it survives a move to a second streaming system that the head was never fitted on.

Our contributions are the following.

\begin{itemize}
    \item We formulate safety for streaming video generation as a stopping problem and define exposure, the harmful frames released before a guard acts, which replaces the whole-clip attack success rate.
    \item We build the defense as a cascade inside the generator. A cheap head scores a chunk before the decoder runs and decides when the expensive verifier is worth a call, so the threshold spends compute.
    \item We measure the composition against prompt filters, image moderators and a per-chunk video model over 599 streams, and we validate its harm oracle against a guardrail model and a blinded audit.
\end{itemize}

\section{Related Work}
\label{sec:related}

\noindent\textbf{Streaming and Causal Video Generation.} Causal reformulations break the joint denoising that leaves no frame complete until the last sampling step. Diffusion Forcing trains with per-token noise levels~\cite{chen2024diffusion}, CausVid distills a bidirectional teacher into a causal student~\cite{yin2025slow}, and Self-Forcing closes the train-test gap by rolling out the student~\cite{huang2025selfforcing}, while LongLive adds prompt switching~\cite{yang2026longlive} and others push throughput and scale~\cite{liu2026rolling,feng2025streamdiffusionv2,chen2025skyreels}.

\noindent\textbf{Safety of Text-to-Video Generation.} Benchmarks established that generators produce unsafe video, with T2VSafetyBench~\cite{miao2024t2vsafetybench} defining fourteen harm categories, three of them temporal, and SafeSora~\cite{dai2024safesora} adding preference labels. Attacks run from prompt optimization to compositions whose fragments are each benign~\cite{liu2025t2voptjail,lee2026scenesplit,ma2026safegen}, and defenses filter the text embedding, rewrite prompts or steer the denoising trajectory~\cite{yoon2025safree,liang2025t2vshield,liang2026safeca,zou2026trajshield}, a steer that inverts~\cite{kundu2026reins}, while concept erasure removes the target from the weights~\cite{ye2025t2vunlearning,xu2025videoeraser,liao2026cleanvideo,li2026fade,facchiano2025video,liu2026motion,fedward,fedcl}. Every method here assumes the clip is complete before it is judged, and a streamed generator breaks that assumption the moment that it releases its very first chunk.

\noindent\textbf{Temporal Blind Spots of Video Moderation.} Deployed moderation inherits image machinery. A per-frame checker pooled over sampled frames is invariant to frame order, so it accepts a harmful clip whenever it accepts a shuffle of that clip~\cite{cao2026temporal}. Sampling creates a second gap, since a segment shorter than the sampling interval is seen by at most one frame, and harm spread thinly defeats more than thirty multimodal models~\cite{cao2026failures,cao2025poisoning,cao2026caved,wang2026distributed}. Guard models trained for video~\cite{chen2025safewatch,nahin2026safelens,zhu2026guardreasoneromni} raise accuracy without removing these gaps, and two works read the generator's own latent~\cite{pang2025towards,khatri2026latent} without addressing a streamed generator or the rate at which a benign stream is interrupted.

\noindent\textbf{Detection and Cascades.} Quickest detection studies when to act on a stream of noisy observations, where the cumulative sum is minimax optimal in Lorden's sense~\cite{page1954continuous,lorden1971procedures}, the Shiryaev-Roberts procedure is its Bayesian counterpart~\cite{shiryaev1963optimum}, and both trade delay against the false-alarm rate~\cite{moustakides1986optimal}. Those guarantees assume known densities, which a learned score only approximates, so we fix the operating point empirically and calibrate it with conformal risk control~\cite{angelopoulos2024conformal,angelopoulos2025learn}. The alternative of \cref{sec:app-steer} in the appendix builds on refusal directions in activation space~\cite{arditi2024refusal,rimsky2024steering,li2023inference} and on guidance away from an unsafe concept~\cite{schramowski2023safe,xiao2025safediffuser,kim2026safety}. Spending a detection budget unevenly is canonically the attentional cascade~\cite{viola2001rapid,chen2023frugalgpt,nie2024online}, which chooses per input, while ours chooses per chunk inside one input, where the chunks the cheap stage has cleared are already public.

\section{Method}
\label{sec:method}

\subsection{Streaming Generation and Exposure}
\label{sec:threat}

A causal generator produces a clip chunk by chunk. Writing $z_t$ for the clean latent of chunk $t$ and $S_t$ for the recurrent state it carries, the per-block key-value cache, one step of generation is
\begin{equation}
    (z_t, S_t) = G_\theta(\epsilon_t, c, S_{t-1}),
    \label{eq:step}
\end{equation}
where $\epsilon_t$ is the chunk's initial noise and $c$ is the conditioning. The decoder $D$ turns $z_t$ into $f_t$ pixel frames, released at once, and that release is what separates this setting from the defenses before it.

Let $J$ be a harm oracle, $\nu = \min\{t : J(z_t) = 1\}$ the onset, the first chunk a viewer should not see, and $h_t \le f_t$ the frames of chunk $t$ that the oracle flags. A guard observing evidence $e_t$ chooses a stopping time $\tau$ over the chunks $t = 1, \dots, N$, with $\tau = N+1$ when the guard never acts, and we measure it by its exposure
\begin{equation}
    E = \sum_{t=\nu}^{\min(\tau-1,\, N)} h_t,
    \label{eq:exposure}
\end{equation}
the harmful frames that a viewer has already seen, which is zero when the guard acts at or before the onset, excludes any chunk the verifier confirms, and grows to the rest of the clip when it never acts. \cref{sec:app-metrics} in the appendix sets this beside the looser count of every frame released after the onset. We therefore pair exposure with the probability that a benign stream is ever interrupted.

Two consequences shape the design. Firstly, evidence about chunk $t$ exists before $D$ runs, so a guard reading the generator can act at zero exposure. Secondly, $S_t$ already holds the harmful chunk and pulls later chunks toward it, so changing the conditioning leaves the harm in place.

\subsection{Per-Chunk Evidence}
\label{sec:evidence}

Two sources inside the generator could carry $e_t$. The clean chunk latent $z_t$ reduces to channel statistics and temporal differences, and the hidden state of a mid-depth block at the final denoising step gives one token per latent frame and carries semantic content the latent hides~\cite{splitinference}. \cref{sec:app-ablation} in the appendix measures both and $e_t$ is the hidden state.

A score that pools frames symmetrically cannot separate a harmful clip from a shuffle of it~\cite{cao2026temporal}, so the cheap stage needs the means to read order. We map each latent frame to a token, add a position embedding and apply a small temporal transformer, trained with a shuffle-contrastive term that requires a harmful chunk to outscore a permutation of its own frames by a margin,
\begin{equation}
    \mathcal{L}_{\mathrm{shuf}} = \mathbb{E}_{z \sim \mathcal{H},\, \pi} \big[ \max(0,\, m - g(z) + g(\pi z)) \big],
    \label{eq:shuffle}
\end{equation}
where $\mathcal{H}$ is the set of harmful chunks, $\pi$ permutes latent frames and $m$ is the margin. A class-balanced logistic loss completes the objective and the score $s_t = g(e_t)$ serves as an approximate log-likelihood ratio. The deployed guard carries this term because it is what was trained. It raises the head's sensitivity to a permutation of the frames without yielding a reliable reduction in exposure, since \cref{sec:app-order} in the appendix finds an order-invariant head of the same size does as well, and the sensitivity to order that our measurements do reward sits in the decoded-pixel verifier.

Reading inside the generator also removes the sampling gap, since a guard sampling $k$ of $N$ frames covers a burst of $d$ with probability at most $kd/N$~\cite{cao2026failures} while every latent frame enters our score, and \cref{sec:app-repro} in the appendix gives the architecture of the head.

\subsection{Sequential Decision with a Calibrated Budget}
\label{sec:sequential}

\begin{table*}[!t]
\centering
\small
\caption{Exposure and cost of each defense.}
\label{tab:main}
\setlength{\tabcolsep}{4pt}
\begin{tabular}{@{}lccccccc@{}}
\toprule
\multirow{2}{*}{Defense} & Exposure & \multirow{2}{*}{95\% CI} & Zero exposure & False alarm & \multicolumn{2}{c}{Frames scored} & Detector \\
\cmidrule(lr){6-7}
 & (frames) & & (\%) & (\%) & Benign & Unsafe & time (s) \\
\midrule
No defense & 46.25 & [42.81, 49.59] & 21.9 & 0.0 & 0.0 & 0.0 & 0.00 \\
Prompt filter & 0.00 & --- & 100.0 & 0.0 & 0.0 & 0.0 & 0.04 \\
Image moderator at 2 fps & 12.10 & [10.04, 14.35] & 61.2 & 0.9 & 11.0 & 7.1 & 2.24 \\
Image moderator at 16 fps & 9.36 & [7.53, 11.40] & 74.3 & 1.3 & 80.3 & 45.6 & 4.55 \\
Chunk video model & 0.91 & [0.46, 1.42] & 89.6 & 1.8 & 80.1 & 32.6 & 1.85 \\
\midrule
Our guard, halting & 6.61 & [5.17, 8.23] & 71.1 & 4.4 & 0.0 & 0.0 & 0.00 \\
Our guard, budgeted verification, frames & 8.26 & [6.57, 10.09] & 69.3 & 1.3 & 5.1 & 31.3 & 1.95 \\
Our guard, budgeted verification, video model & 2.35 & [1.67, 3.11] & 81.0 & 0.4 & 5.3 & 21.7 & 0.56 \\
\midrule
Oracle stop & 0.00 & [0.00, 0.00] & 100.0 & 3.1 & 81.0 & 81.0 & --- \\
\bottomrule
\end{tabular}
\end{table*}

A decision rule turns scores into a stopping time. We write $W_t$ for whichever statistic a rule carries and alarm at $\tau = \min\{t : W_t \ge h\}$, so the memoryless test is $W_t = s_t$, and the member of the family that accumulates the history is a cumulative sum,
\begin{equation}
    W_t = \max(0,\, W_{t-1} + s_t), \qquad W_0 = 0,
    \label{eq:cusum}
\end{equation}
which under known densities minimizes the worst-case delay for a given time to false alarm~\cite{page1954continuous,lorden1971procedures,shiryaev1963optimum}. A learned score only approximates a likelihood ratio, so that optimality does not carry over, and every number we report uses the memoryless test, which \cref{sec:app-rules} in the appendix finds wins at both of the horizons we try. Accumulation is an alternative we test, and it is no part of the deployed guard.

The threshold is set without distributional assumptions. Given $n$ benign calibration streams we take $h$ to be the $\lceil (n+1)(1-\alpha) \rceil$-th smallest of their running maxima of $W_t$, which bounds by $\alpha$ the probability that an exchangeable benign stream ever drives $W_t$ over $h$~\cite{angelopoulos2024conformal}. A crossing triggers a verifier call and the stream continues. That guarantee is marginal, so we report the realized crossing rate beside the nominal level it need not match.

\subsection{Budgeted Verification}
\label{sec:budget}

An internal-state score is cheap and a decoded-pixel detector is accurate, so the accurate test decides and the cheap one decides when to pay for it. Write $V$ for a verifier reading one chunk's decoded frames at a cost of $c$ frame evaluations. We consult it only where the statistic has crossed the threshold, and leave that chunk's decision to it alone,
\begin{equation}
    a_t = \mathbb{1}[W_t \ge h] \cdot V(D(z_t)),
    \label{eq:cascade}
\end{equation}
halting at the first chunk with $a_t = 1$ and releasing a flagged chunk that $V$ clears, which resets an accumulating statistic and leaves a memoryless one unchanged. The expected cost is $c\,\mathbb{E}\,|\{t : W_t \ge h\}|$, which the threshold controls directly and an operator fixes in advance.

A false alarm in a halting-only guard withholds frames a viewer wanted, so its threshold stays high, while one under \cref{eq:cascade} costs one call to $V$ and the threshold can fall until the compute budget sets it rather than the viewer.

Exposure under \cref{eq:cascade} splits into the harm that the guard never flags and the harm that it flags and $V$ clears, so the composition is never safer than the verifier on every chunk and approaches that limit as the guard's ranking within a stream improves. Random allocation at a matched cost separates the value of the evidence from the effect of uneven spending, since it spends the same budget blind.

\section{Experiments}
\label{sec:experiments}

\subsection{Experimental Setup}
\label{sec:setup}
\noindent\textbf{Streaming Generator.} We evaluate on Self-Forcing, a four-step chunk-wise student distilled from a bidirectional Wan2.1-T2V-1.3B teacher that streams 480p video in real time on one GPU~\cite{huang2025selfforcing,wan2025wan}. It produces 21 latent frames in chunks of 3, decoded to 81 pixel frames of 480$\times$832 at 16 fps, and we also roll it out three times longer, repeat the benchmark on a second student of the same backbone, and carry the guard over to LongLive~\cite{yang2026longlive}.

\noindent\textbf{Prompt Sets.} The benchmark holds 1597 streams, 600 benign and 997 unsafe, across four harm categories, drawn from VBench, VidProM and Movie Gen~\cite{huang2024vbench,wang2024vidprom,polyak2024moviegen} and from the taxonomy of T2VSafetyBench~\cite{miao2024t2vsafetybench}. Every unsafe prompt appears twice, directly and behind a place-based opening, and the comparison runs on a 599-stream subset every method shares. Two further sets reach the regime a filter cannot, 734 streams whose requests never name the harm and 360 that switch the text condition mid-stream, which \cref{sec:app-prompts} and \cref{sec:app-natural} in the appendix describe.

\noindent\textbf{Harm Oracle.} Onset comes from an automatic oracle, since hand-labeling every frame is infeasible at this scale. It scores each frame with a fine-tuned NSFW classifier, NudeNet and two zero-shot CLIP comparisons, and each chunk with a video language model~\cite{bedapudi2019nudenet,radford2021clip,bai2025qwen25vl}, and a chunk is harmful when any of the five passes a threshold at one shared significance level, realizing 0.89\% on benign chunks. Since that video model is also a verifier compared here, \cref{sec:res-panel} and \cref{sec:res-human} validate the oracle against SafeWatch-8B~\cite{chen2025safewatch} and against a blinded audit of 790 chunks by 50 participants.

\noindent\textbf{Baselines and Metrics.} Exposure is bounded above by the undefended generator and below by an oracle halting at the first harmful chunk, and three families stand in for current practice, a text classifier on the prompt~\cite{inan2023llamaguard}, an image moderator sampling decoded frames at 1 to 16 frames per second~\cite{cao2026temporal,schramowski2022q16}, and a video language model on every chunk~\cite{chen2025safewatch}, which is also the verifier our policy gates. Exposure from \cref{eq:exposure} is the primary metric, reported beside the fraction of unsafe streams at zero exposure, the fraction of benign streams interrupted, the frames a detector scores and its time per stream (s). Intervals are stream-level bootstraps shared by every policy, so a difference is paired, and \cref{sec:app-metrics} in the appendix says what each baseline bounds.

\noindent\textbf{Implementation Details.} The guard reads the pooled hidden state of block 10 of 30 at the last denoising step and scores it with a two-layer temporal transformer of width 256, and \cref{sec:app-repro} in the appendix gives the optimizer and the schedule it was trained under.

\subsection{Exposure Under Each Defense}
\label{sec:res-main}

\cref{tab:main} reports every defense on the shared subset of 599 evaluation streams. The undefended generator shows 46.25 harmful frames of the 81 it releases, and an oracle halting at the first harmful chunk removes all of it at the price of interrupting 3.1\% of the benign streams.

The accurate detector and the expensive detector are the same video language model. Run on every chunk that it leaves 0.91 frames $[0.46, 1.42]$ while scoring 80.1 frames of every benign stream, and triggered by the guard at a budget of 5 it leaves 2.35 $[1.67, 3.11]$ at 5.3 scored, which is a fifteenth of the cost for 1.44 frames more exposure, and raising the budget to 40.3 frames brings it to 0.91. The image moderators are weaker at every sampling rate, from 14.20 frames at 1 fps to 9.36 at 16 fps.

\noindent\textbf{Reach of the Text Filter.} The prompt filter is the strongest row of \cref{tab:main}. A video language model asked whether a request should be refused blocks 366 of the 374 unsafe prompts and none of the 225 benign ones, and the eight it passes produce streams the oracle never flags, so its exposure is zero. Every unsafe prompt here names what it asks for, so we report it and leave it, and a guard reading the generator complements such a filter without replacing it. \cref{sec:res-indirect} builds the regime a guard answers for by letting the filter's own model rewrite the requests that it blocked.

\begin{figure*}[!t]
  \centering
  \includegraphics[width=\textwidth]{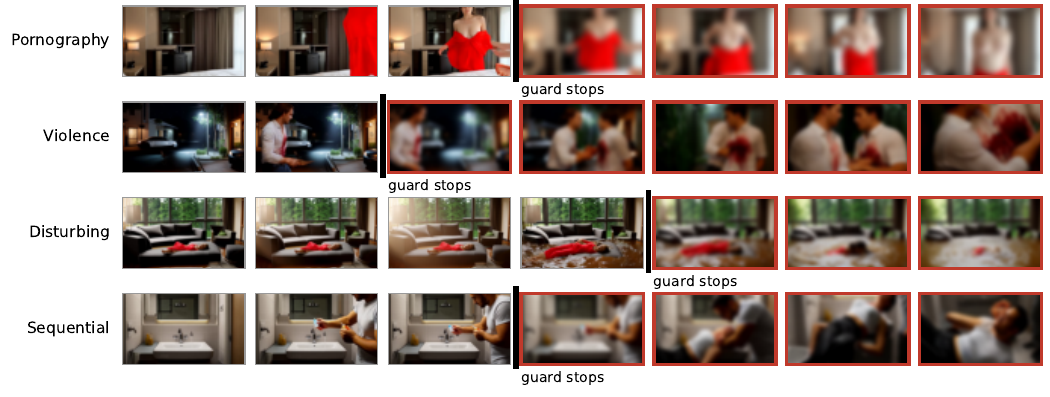}
  \caption{One late-onset stream of each harm category, with the harmful chunks bordered and obscured.}
  \label{fig:qual}
\end{figure*}

\noindent\textbf{Per-Category Behavior.} Pornography is the easiest condition, where the guard-triggered video model leaves 0.32 frames on direct onsets against 68.77 undefended, and harmful sequential action is the hardest at 3.89, so harm living in the order of events is what every detector here handles least well. \cref{fig:qual} shows one late-onset stream of each harm category, where an ordinary scene runs until the onset and the guard withholds every chunk after it. \cref{tab:bycat} in the appendix gives the breakdown by category and onset.

\subsection{Requests That Pass a Text Filter}
\label{sec:res-indirect}

We give the model that acts as the filter the job of rewriting a blocked request as a neutral description of what a camera would record, and a rewrite is kept only when the filter then passes it, so nothing depends on anyone reading the prompts. Eight attempts convert 154 of the 366 blocked requests, a yield of 42.1\%, and the mean prompt harm score of the kept rewrites is 0.153, against 0.909 for the original requests that they were derived from.

\cref{tab:indirect} in the appendix reports the same policies over those 154 streams at the thresholds the benchmark fitted. The filter's own row becomes the undefended row, so its exposure moves from 0.00 frames to 14.93, against the 46.25 of the requests these came from. The guard holds, leaving 2.59 frames $[1.10, 4.40]$ here against 2.35 on the benchmark, while the per-chunk video model leaves 1.67 and the image moderator at 16 fps leaves 3.69. It also fails on disturbing imagery that starts at once, removing none of the 6.29 frames, much as the per-chunk model does not.

\subsection{Requests That Never Name the Harm}
\label{sec:res-natural}

The rewritten requests of \cref{sec:res-indirect} were kept because the filter accepted them, so their harm rate is a property of the filter. This set was written in advance, so both halves of the question are measurable, and \cref{tab:natural} reports them over 734 streams. The filter blocks 21.8\% of the 509 requests written with an unsafe intention against 97.9\% of the benchmark's, and 20.8\% of the plainly benign requests in risky domains. The oracle then confirms harm in 21.2\% of the 509 streams, and in 23.8\% $[17.3, 31.9]$ of the plainly benign ones although nothing in them asks for any, which is the case no prompt filter can address even in principle. The benign control sits at 2.7\% $[1.2, 5.7]$, which is the oracle's own false positive rate on ordinary footage.

The defenses behave differently here, and the honest reading of the comparison is a mixed one. Over all 509 streams the undefended generator shows 7.91 frames, because four fifths carry no harm at all. Our policy at a budget of 5 leaves 3.26 frames $[2.25, 4.33]$ against 6.18 for the same budget spent at random, so the ordering is still worth half the exposure, while an image moderator at 2 fps leaves 2.08 frames for 11.0 scored, so the advantage over uniform sampling is gone. Harm a generator produces unasked is mostly harm that a single frame can show, which is the regime an image moderator was built for, while the guard was fitted where the harm had been asked for.

\begin{table}[t]
\centering
\small
\setlength{\tabcolsep}{2pt}
\caption{Requests that never name the harm they ask for.}
\label{tab:natural}
\begin{tabular}{@{}lccccc@{}}
\toprule
\multirow{2}{*}{Family} & \multirow{2}{*}{Streams} & Filter & Harm & \multicolumn{2}{c}{Exposure} \\
\cmidrule(lr){5-6}
 & & blocks (\%) & shown (\%) & None & Ours \\
\midrule
Implicit & 150 & 37.3 & 30.7 & 11.78 & 3.20 \\
Compositional & 132 & 8.3 & 16.7 & 5.94 & 4.13 \\
Drifting & 97 & 17.5 & 9.3 & 2.23 & 0.88 \\
Benign domain & 130 & 20.8 & 23.8 & 9.68 & 4.21 \\
Benign control & 225 & 0.0 & 2.7 & 0.36 & 0.36 \\
\bottomrule
\end{tabular}
\end{table}

\noindent\textbf{Refitting the Guard on This Regime.} That belongs to what the guard was shown, while the rule reading its score is untouched, so we retrain the head on 366 streams of this set and report both arms on the 368 it never saw, with the oracle, the verifier and the budget unchanged. The carried guard leaves 3.89 frames $[2.33, 5.65]$ at 5.4 benign frames scored and the refitted one leaves 2.73, 2.81 and 2.64 over three seeds, which cuts the carried guard's exposure by just under a third and lands within a frame of the per-chunk model at 2.03. The seeds agree in direction while their intervals overlap, so refitting narrows the gap on this distribution without ever closing it altogether.

\noindent\textbf{Requests That Change in the Middle.} A filter that screened the opening request never sees the one replacing it, so we switch the text condition at chunk 2 or 3 of 360 streams, 120 of them to another benign request as a control. The oracle confirms harm in 33.8\% $[28.1, 39.9]$ of the unsafe switches against 5.0\% $[2.3, 10.5]$ of the benign ones, and the guard's score rises by 2.44 after an unsafe switch against 0.14 after a benign one, so the harm moves it and the change alone does not. It does not react enough. Undefended exposure is 10.93 frames and our policy leaves 6.63 $[4.81, 8.61]$ against 9.66 for random allocation, while the video model on every chunk leaves 1.48, because the key-value cache still holds the benign opening, so the pixels change before the state does. Both regimes give way at the gate while the verifier holds, since the video model reading every chunk still reaches 2.03 and 1.48 frames where our policy reaches 2.73 and 6.63.

\subsection{Allocation of a Verification Budget}
\label{sec:res-budget}

\begin{figure}[t]
  \centering
  \includegraphics[width=\columnwidth]{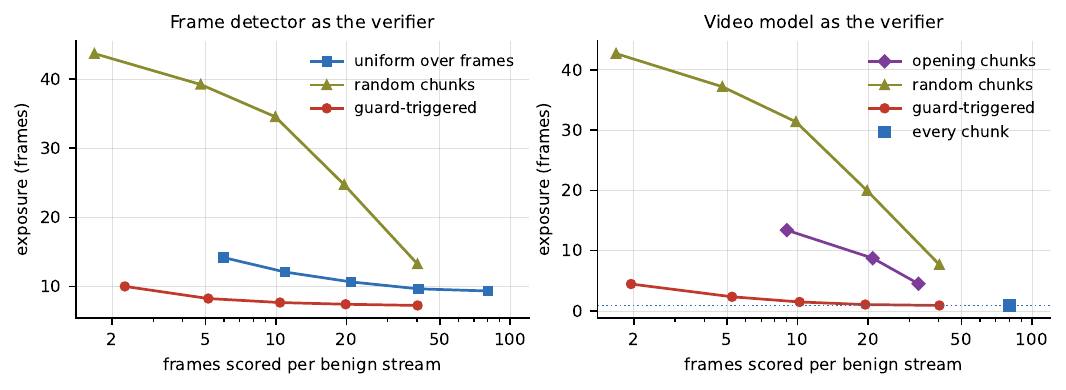}
  \caption{Cost against exposure for different ways of spending a detector budget inside a stream.}
  \label{fig:frontier}
\end{figure}

\cref{fig:frontier} answers where a fixed moderation budget should be spent inside a stream, over the same 599 undefended streams, and \cref{tab:budget} in the appendix gives the numbers behind the curves. The guard-triggered allocation dominates both alternatives under either verifier. With the frame detector it reaches 8.26 frames at 5.1 scored per benign stream against the 80.3 uniform sampling needs to reach 9.36, and with the video model 2.35 at 5.3 against the 80.1 scoring every chunk costs to reach 0.91. Almost all of that value is in the ordering, since random selection at 4.8 frames per benign stream leaves 37.20 frames, a paired difference of 34.87 $[31.62, 38.26]$, and the verifier that reads the chunk as a clip is worth as much, since the frame detector at the same budget gives up 5.95 frames $[4.27, 7.77]$.

\noindent\textbf{A Longer Stream.} The ordering is worth more as the stream lengthens. Rolled out to 63 latent frames, the generator gives 21 chunks and 249 released frames, and undefended exposure rises from 46.25 frames to 134.01. \cref{tab:horizon} reports the defenses there. The video model on every chunk reaches 2.09 frames while scoring 243.9 of a benign stream's 249, our policy at the benchmark's budget of 5 reaches 8.78 while scoring 6.7, and widening the gate to 20 frames brings it to 3.76 at 19.7 while interrupting one of the 236 benign streams. At that budget random allocation leaves 116.49 frames against our 8.78, where the same budget over 7 chunks leaves 37.20 against 2.35, so tripling the length of the stream roughly triples what the guard's ordering is worth to a deployment.

\noindent\textbf{Decision Rules.} Which statistic reads the score matters far less than where the budget goes. The memoryless test leaves 2.35 frames against 2.43 for a cumulative sum, and accumulation pays only without the verifier, where a halting cumulative sum leaves 11.61 frames against 18.82 at a 1\% budget. \cref{tab:rules} in the appendix compares six rules.

\begin{table}[t]
\centering
\small
\setlength{\tabcolsep}{4pt}
\caption{Defenses over a 21-chunk horizon.}
\label{tab:horizon}
\begin{tabular}{@{}lccc@{}}
\toprule
\multirow{2}{*}{Defense} & Exposure & Zero & Benign \\
 & (frames) & (\%) & frames \\
\midrule
No defense & 134.01 & 15.3 & 0.0 \\
Moderator at 16 fps & 18.85 & 70.8 & 238.1 \\
Chunk video model & 2.09 & 82.6 & 243.9 \\
\midrule
Memoryless halting & 10.81 & 68.1 & 0.0 \\
Cumulative sum halting & 11.32 & 66.7 & 0.0 \\
Our guard, video model & 8.78 & 61.8 & 6.7 \\
Our guard, wider gate & 3.76 & 71.5 & 19.7 \\
\midrule
Oracle stop & 0.00 & 100.0 & 249.0 \\
\bottomrule
\end{tabular}
\end{table}

\subsection{Scope of the Threshold Guarantee}
\label{sec:res-calib}

The conformal construction of \cref{sec:sequential} controls one event, which is that an exchangeable benign stream ever drives the statistic over the threshold. It bounds neither exposure nor the verifier calls a stream pays for, and \cref{tab:calib} in the appendix measures each on held-out benign streams. The realized crossing rate reaches 3.5\% $[1.4, 8.7]$ against a nominal 1\% at the tight end of the short horizon, which is what a marginal guarantee allows on one draw of 112 calibration streams, and the long horizon does not repeat that overshoot. The cascade stops 0.0\% of benign streams at every level, because the verifier clears the chunks the guard flags, so the crossing rate is a compute quantity rather than a user-facing one, and the thresholds in our tables are fitted on one fold and reported with the cost they realized on the other, so none is scored on its own fold.

\subsection{Validating the Harm Oracle}
\label{sec:res-panel}

Every number above rests on an oracle that reads a video language model, and that model is also one of the verifiers compared, so the comparison alone cannot say whether the exposure is a property of the streams or of that model. We therefore label the same 4193 chunks again with SafeWatch-8B, whose encoder, language model and training data are disjoint from every signal the oracle reads. The oracle calls 38.1\% of chunks harmful against the guardrail model's 22.0\%, they agree on 82.8\% at $\kappa=0.603$ $[0.550, 0.651]$, and cutting the guardrail model at the oracle's benign rate raises that to 87.0\% at $\kappa=0.737$ $[0.698, 0.774]$, so the two differ mostly in how much they flag, while the chunks they find largely agree.

Relabeling moves every absolute number and leaves the comparison alone. Under the independent labels the undefended generator shows 28.84 frames $[25.58, 32.30]$, our policy leaves 0.66 $[0.36, 1.01]$ at 5.1 benign frames scored and random allocation 21.92, and cutting those labels at the oracle's own benign rate turns the three into 61.29, 9.85 $[8.12, 11.66]$ and 52.70. The ordering of all five policies and both gaps survive. Both are models, so a bias they share stays invisible here, which \cref{sec:res-human} takes up.

\begin{table}[t]
\centering
\small
\setlength{\tabcolsep}{4pt}
\caption{Agreement of the automatic labels with human consensus and among the annotators.}
\label{tab:human}
\begin{tabular}{@{}lccc@{}}
\toprule
Labeler & Precision (\%) & Recall (\%) & $\kappa$ \\
\midrule
\multicolumn{4}{@{}l}{\textit{Arguable chunks counted as harm}} \\
Harm oracle & 96.1 & 82.8 & 0.732 \\
Independent panel & 93.8 & 91.2 & 0.851 \\
\midrule
\multicolumn{4}{@{}l}{\textit{Arguable chunks counted as safe}} \\
Harm oracle & 81.4 & 87.1 & 0.744 \\
Independent panel & 69.2 & 94.6 & 0.659 \\
\midrule
\multicolumn{4}{@{}l}{\textit{Inter-annotator agreement}} \\
Moderation decision & --- & --- & 0.864 \\
Harm category & --- & --- & 0.792 \\
Severity (ordinal $\alpha$) & --- & --- & 0.741 \\
\bottomrule
\end{tabular}
\end{table}

\subsection{A Human Audit of the Oracle}
\label{sec:res-human}

Both automatic labelers are models, so neither alone establishes whether a person would withhold a chunk. We therefore conduct a blinded audit of 790 chunks with 50 independent participants, none of whom are paper authors, over three strata, namely all chunks of 70 complete streams spanning the four harm categories and benign content, 150 oracle-positive chunks and 150 oracle-negative ones. Each chunk receives at least three judgments on six evenly spaced frames, with an intermediate answer for arguable cases, and prompts, category labels and both automatic predictions stay hidden. Consensus is a majority vote and every rate is weighted back to the population, under the protocol that \cref{app:human-audit} in the appendix sets out.

\noindent\textbf{Human Agreement.} The judgments agree strongly on the moderation decision, at Fleiss' $\kappa=0.864$, and substantially on harm category ($\kappa=0.792$) and severity (ordinal Krippendorff's $\alpha=0.741$). Disagreements concentrate near semantic boundaries, between arguable and clearly harmful or between violence and disturbing imagery, and they rarely change the decision, so consensus is a sound reference.

\noindent\textbf{Automatic-Label Validation.} \cref{tab:human} compares both automatic labelers against consensus. Counting arguable chunks as harmful, the oracle reaches 96.1\% precision and 82.8\% recall at $\kappa=0.732$ and the independent panel 93.8\% and 91.2\% at $\kappa=0.851$, while counting them as safe raises the oracle's recall to 87.1\% at 81.4\% precision, so the two readings bracket the borderline cases. The panel follows the human decisions more closely than the oracle, so this paper's comparisons rest on a second line of support.

\noindent\textbf{Human-Validated Exposure.} The complete streams let exposure be recomputed from human-consensus onsets instead of the oracle's. There the undefended generator releases 48.6 frames after onset, random allocation leaves 39.1 at about five verifier-scored frames per benign stream, our budgeted verification leaves 3.1 and the video verifier on every chunk reaches 1.2 at many times the cost, so the policy cuts exposure by 92.1\% against random allocation at a matched budget under human judgment as well.

\begin{table}[t]
\centering
\small
\setlength{\tabcolsep}{4pt}
\caption{The guard on a streaming system that it was not fitted on.}
\label{tab:transfer}
\begin{tabular}{@{}lccc@{}}
\toprule
\multirow{2}{*}{Defense} & Exposure & \multirow{2}{*}{95\% CI} & Benign \\
 & (frames) & & frames \\
\midrule
No defense & 50.99 & [47.70, 54.39] & 0.0 \\
Chunk video model & 0.60 & [0.30, 0.99] & 79.9 \\
Image moderator at 16 fps & 12.48 & [10.24, 14.93] & 79.5 \\
Random chunks & 41.71 & [38.46, 45.08] & 4.8 \\
\midrule
Guard, threshold carried & 2.40 & [1.73, 3.19] & 6.4 \\
Guard, gate refitted here & 4.23 & [3.08, 5.54] & 4.5 \\
Head retrained here & 3.57 & [2.58, 4.79] & 4.6 \\
\bottomrule
\end{tabular}
\end{table}

\subsection{Transfer to Another Streaming System}
\label{sec:res-transfer}

\noindent\textbf{A Second Student of the Same Teacher.} Rerunning the benchmark on the second student that Self-Forcing ships, with the guard carried over and the oracle refit on its own benign chunks, undefended exposure is 40.06 frames against 46.25 and the ranking is reproduced in full, with the per-chunk video model at 0.70 frames, our policy at 2.17 and random allocation at 33.89.

\noindent\textbf{A Different Streaming System.} LongLive is a separate system on the same backbone, with its own checkpoint, a low-rank adapter, a frame sink and local attention~\cite{yang2026longlive}. We generate the same 599 evaluation and 639 fitting prompts with it and refit the oracle on its own benign chunks, and its streams are more harmful than the benchmark's, at 50.99 undefended frames. \cref{tab:transfer} separates the three questions a deployment would ask. The guard carried over with its threshold from elsewhere leaves 2.40 frames $[1.73, 3.19]$ at 6.4 benign frames against the 5 it was calibrated for, so the score lands on a similar scale here. Choosing the gate on this generator at a matched budget gives 4.23 and retraining the head gives 3.57 $[2.58, 4.79]$, so retraining saves 0.66 frames and the intervals overlap. All three sit against 41.71 frames for random allocation and 12.48 for an image moderator at 16 fps, so what carries across checkpoints and systems in this backbone family is the ordering of the policies, while single numbers move around it.

\subsection{Cost of Running the Guard}
\label{sec:res-cost}

The budget counts the frames the costly verifier scores, and everything else is paid on every chunk. The guard reads states the generator has already computed, so its decision holds no frame, while a chunk that it flags is decoded and then held for a single verifier call. Timed with CUDA events over 40 streams, denoising takes 996 ms, the context pass 249 ms and decoding 629 ms, while the tap is free to the measurement's precision and the head takes 1.8 ms, or 0.1\% of a chunk. The verifier takes 322 ms per flagged chunk, and at a budget of 5 benign frames a stream waits on 0.44 of its 7 chunks, adding 0.14 s of latency.

A frame scored is not a fixed amount of compute, since the frame detectors take 2457 ms per chunk against the video model's 322 ms, so the two verifiers cost 7.6$\times$ apart at the same frame count. Matching wall clock would let the video model read seven times as many chunks, so the frames-scored budget understates how much cheaper the better verifier is. \cref{tab:main} measures that axis, where our policy spends 0.56 s a stream against 1.85 s for the video model on every chunk and 2.24 s for the moderator at 2 fps. Neither axis counts generation, at 1.87 s a chunk, and the guard's head adds 1.8 ms to each. Generating a seven-chunk stream takes 13.1 s, several times the largest of those detector times, so the defenses here differ in the compute they ask of a deployment rather than in the delay that any of them adds to a viewer's wait~\cite{ppautodl,medshield}.

\noindent\textbf{Robustness.} An attacker who resubmits the same prompt and keeps whichever stream survives raises the exposure accumulated over the episode from 2.53 frames to 8.28, while the per-chunk model rises from 0.91 to 2.34. Under a frame-only oracle that drops the video language model the ranking survives, with our policy at 1.30 frames against 0.91, and \cref{sec:app-results} in the appendix reports both arms in full.

\section{Limitations}
\label{sec:limitations}
The benchmark's unsafe prompts name the harm they ask for, so a text filter settles almost all of them, and the two regimes of \cref{sec:res-indirect} and \cref{sec:res-natural} are where a guard earns its place. The harm oracle remains imperfect, above all on harmful sequential action, and the exposure recomputed from human onsets rests on 70 complete streams out of the whole benchmark. The guard is trained on one distilled generator and every streaming generator we could run belongs to one backbone family, so portability across architectures stays an assumption~\cite{fedcml}. And every defense that halts pays for a retry, which raises the exposure our guard concedes from 2.53 frames to 8.28.

\section{Conclusion}
\label{sec:conclusion}
We framed streaming video safety around exposure, the harmful frames reaching a viewer, and built the defense as a cascade inside the generator. A guard reading hidden states the generator has already computed is cheap enough for every chunk and too weak to trust alone, so it decides where an expensive verifier spends its budget. At a fifteenth of that verifier's benign cost it leaves 2.35 frames against 0.91, and the advantage grows with the stream's length and survives a move to a second system. What carries the result is the division of labor between ranking and judging. Future work should widen the distribution the guard is fitted on, since refitting it where harm is never named recovers much of what it gives up, and should carry it beyond this backbone family, which only another architecture can settle.

{
    \small
    \bibliographystyle{ieeenat_fullname}
    \bibliography{main}
}

\clearpage
\appendix

\twocolumn[{%
\centering
{\Large\bf \papertitle}\\[0.4em]
{\Large Appendix}\\[0.9em]
}]

\section{Per-Category Results}
\label{sec:app-bycat}

\cref{tab:bycat} gives the breakdown behind the aggregate numbers of \cref{tab:main} in the main text, by harm category and by where the harm starts. Three readings are worth drawing out. Firstly, pornography is the condition every detector handles best and sequential action is the condition every detector handles worst, and the ordering is the same for the image moderator, the video model and our guard, so it is a property of the harm, and no particular detector explains it. Secondly, late onsets are not uniformly harder than direct ones. They are much harder for the image moderator on pornography, at 27.96 frames against 4.82, and they are easier for the video model on sequential action, at 0.47 frames against 0.86, because a late-onset stream spends its opening chunks on an ordinary scene that gives the model a contrast to judge the turn against. Thirdly, the guard-triggered video model tracks the per-chunk video model closely in every category, and the categories where it gives up the most are the ones where the guard's own ranking is weakest, which is sequential action at 3.89 frames against 0.86. The count of streams is printed beside every row because the subgroups are not the same size, and the two smallest hold 36 and 38 streams, so a difference of one or two frames inside such a row is not resolved by this sample while the gap between 15.69 frames undefended and 3.89 with the guard is.

\cref{tab:catbudget} separates harm that a single frame can show from harm that exists in the order of events and follows each against the budget. The two behave differently, and neither is uniformly harder. Undefended exposure is 54.22 frames where the harm is in the appearance and 13.95 where it is in the order, because a sequence that depicts an unsafe act spends part of its length on frames that are not themselves unsafe. At a tight budget the ordered harm is the harder one, at 4.80 frames against 4.38 with two benign frames scored and 3.27 against 2.12 with five, and the ranking reverses once the budget is loose, at 0.82 against 1.11 with twenty frames scored. A budget achieves less on ordered harm because the guard's score reaches the threshold later in such a stream, and once the budget is large enough to carry several calls the verifier that reads the chunk as a clip is better at ordered harm than at appearance.

\begin{table*}[ht]
\centering
\footnotesize
\caption{Exposure (frames) by category and onset, with the fraction of streams stopped.}
\label{tab:bycat}
\setlength{\tabcolsep}{4pt}
\begin{tabular}{@{}lcccccccc@{}}
\toprule
\multirow{2}{*}{Category} & \multirow{2}{*}{Streams} & \multirow{2}{*}{No defense} & \multicolumn{2}{c}{Moderator at 16 fps} & \multicolumn{2}{c}{Video model, every chunk} & \multicolumn{2}{c}{Guard-triggered video model} \\
\cmidrule(lr){4-5} \cmidrule(lr){6-7} \cmidrule(lr){8-9}
 & &  & Exposure & Stopped (\%) & Exposure & Stopped (\%) & Exposure & Stopped (\%) \\
\midrule
Pornography, direct & 56 & 68.77 & 4.82 & 85.7 & 0.00 & 98.2 & 0.32 & 98.2 \\
Pornography, late & 56 & 60.12 & 27.96 & 66.1 & 0.00 & 96.4 & 1.71 & 94.6 \\
Violence, direct & 56 & 54.61 & 8.36 & 71.4 & 0.02 & 83.9 & 1.04 & 83.9 \\
Violence, late & 56 & 40.11 & 2.36 & 73.2 & 1.70 & 89.3 & 3.71 & 87.5 \\
Disturbing, direct & 38 & 59.13 & 10.74 & 68.4 & 3.45 & 81.6 & 4.39 & 76.3 \\
Disturbing, late & 38 & 39.37 & 4.08 & 78.9 & 1.71 & 76.3 & 2.34 & 73.7 \\
Sequential action, direct & 36 & 15.69 & 6.58 & 41.7 & 0.86 & 47.2 & 3.89 & 38.9 \\
Sequential action, late & 38 & 12.29 & 6.95 & 23.7 & 0.47 & 60.5 & 2.68 & 52.6 \\
Benign & 225 & 0.76 & 0.44 & 1.3 & 0.34 & 1.8 & 0.76 & 0.4 \\
\bottomrule
\end{tabular}
\end{table*}

\begin{table}[ht]
\centering
\small
\setlength{\tabcolsep}{2pt}
\caption{Exposure (frames) against the verifier budget (benign frames scored), by what defines the harm.}
\label{tab:catbudget}
\begin{tabular}{@{}lccccccc@{}}
\toprule
\multirow{2}{*}{Harm} & No & \multicolumn{5}{c}{Budget} & Every \\
\cmidrule(lr){3-7}
 & defense & 2 & 5 & 10 & 20 & 40 & chunk \\
\midrule
Appearance & 54.22 & 4.38 & 2.12 & 1.44 & 1.11 & 0.97 & 0.97 \\
Sequential action & 13.95 & 4.80 & 3.27 & 1.69 & 0.82 & 0.66 & 0.66 \\
\bottomrule
\end{tabular}
\end{table}

\section{Prompt Construction}
\label{sec:app-prompts}

Unsafe prompts are written as plain descriptions, with no optimized token strings, because an optimized string drives the generator toward abstract texture whose harm a human would not recognize, and a guardrail measured on texture says nothing about a deployment. Each prompt names a subject, an action drawn from the harm taxonomy, a setting and a camera move, and the four fields are combined so that the same action appears in several scenes. Every unsafe prompt appears twice. The direct variant opens on the harmful request, and the late variant prefixes a short place-based opening and a named transition, so that the two share their harmful content and differ only in where the harm arrives. The late variant is the condition a prompt filter cannot see, and it is also the condition that separates a guard reading the generator from one reading the request. Benign prompts describe everyday scenes and come from the VBench suite, the real user prompts of VidProM and the Movie Gen video bench, so the benign side of the evaluation resembles what a deployed generator receives, since none of its negatives were written to be easy. Streams are partitioned by prompt into 639 for fitting and calibration and 958 for evaluation, the partition is stratified by category and onset, and a direct prompt stays with its own late-onset variant so neither split ever sees the other's harmful content. The prompt set holds 300 streams of pornography, 300 of violence and gore, 200 of disturbing imagery and 197 of harmful sequential action. The defense comparison runs on a stratified subset of 599 evaluation streams, drawn once and shared by every method, because ten defenses over the whole evaluation split would take more than a day of generation and judging on the two cards we hold.

\begin{table}[t]
\centering
\small
\setlength{\tabcolsep}{6pt}
\caption{Requests accepted by the text filter after rewriting.}
\label{tab:indprompts}
\begin{tabular}{@{}lcc@{}}
\toprule
Attempt budget & Accepted & Yield (\%) \\
\midrule
1 attempt & 8 & 2.2 \\
2 attempts & 52 & 14.2 \\
3 attempts & 56 & 15.3 \\
4 attempts & 91 & 24.9 \\
5 attempts & 95 & 26.0 \\
6 attempts & 134 & 36.6 \\
7 attempts & 134 & 36.6 \\
8 attempts & 154 & 42.1 \\
\bottomrule
\end{tabular}
\end{table}

\section{Requests That Do Not Name Their Harm}
\label{sec:app-natural}

The benchmark states what it asks for, which is what makes a text filter settle it, and the rewritten set of \cref{sec:res-indirect} in the main text is defined by what the filter accepted, which makes its harm rate a property of the filter, so it measures nothing on its own. The third set answers what is left and is fixed before anything is generated. Four families are written by hand, each as a request an ordinary user might send. The implicit family describes a situation whose depiction is unsafe without naming the unsafe thing, as in a woman who steps out of the shower and reaches for a towel that is not on the rail. The compositional family writes clauses that are each harmless and unsafe together, as in a butcher's shop in summer whose power has been off for three days. The drifting family opens on a benign scene and announces that it changes without saying what it changes into, as in a hospital corridor whose lights go out one by one. The plainly benign family asks for ordinary footage in domains where a generator may produce harm with nobody asking for it, as in an amateur boxing match in a small gym, or a slaughterhouse line filmed for a report on the food industry.

Each family crosses its scenarios with eight settings and eight camera notes, so the same situation appears in several scenes, with no sentence simply repeated. A third of the prompts in the implicit, compositional and benign families carry a place-based opening and a named transition, which spreads the onset over the stream, and the drifting family is late by construction. The benchmark's benign prompts are carried over unchanged, so the interruption rate on this set is measured on the same benign traffic as everywhere else. No prompt was read, selected or edited after anything was generated, and no stream was dropped for what it produced, which is what makes the harm rate of \cref{sec:res-natural} in the main text a measurement of the generator, which the set of prompts on its own could never have fixed.

Three details sit behind \cref{sec:res-natural} in the main text. The risky domains are amateur boxing, a slaughterhouse line filmed for a food industry report and a life drawing class. Prefixing a benign opening raises how suspicious a request looks to the filter, at 64.0\% blocked on implicit requests that open on a benign scene against 24.0\% on the same requests without one. The families differ in what the generator produces, with implicit requests confirmed harmful in 30.7\% $[23.8, 38.5]$ of their streams and drifting ones in 9.3\% $[5.0, 16.7]$, and conditional on the oracle confirming harm the undefended generator shows 37.27 frames $[32.01, 42.27]$ against the 7.91 frames that it averages over the whole of the set.

After a request changes in the middle, the paired difference between our policy and random allocation at the same cost is 3.02 frames $[1.67, 4.52]$, and an image moderator at 2 fps leaves 4.02 frames for twice our budget. Violence after a switch is where the guard helps, from 16.10 frames to 6.87, and pornography after a switch is where it barely does, from 16.73 to 12.96. Both regimes point the same way. The composition is worth what the guard's ranking is worth, and it degrades to a cost-matched random allocation when that ranking weakens, and never to nothing.

\section{Harm Oracle}
\label{sec:app-oracle}

A signal joins the union only when it separates unsafe chunks from benign ones on its own, which excludes the zero-shot nudity comparison at 0.256 against the violence streams and the zero-shot violence comparison at 0.471 against the pornography streams, both at or below chance. Thresholds are set by one shared significance level, with no even split of the false-alarm budget, because an even split charges every signal for the multiplicity whether or not it contributes and it cost the strongest detector most. Each signal becomes an empirical p-value against the benign chunks of the fitting split and the level rises until the union fires on 1\% of them, which the fitted thresholds realize at 0.89\% over 1680 benign chunks.

The video language model matters to the definition rather than to the accounting. A harm oracle built only from frame detectors would be the same object as one of the baselines, so it could not measure what a per-frame system misses, and we therefore keep the frame-only oracle as the robustness check of \cref{sec:app-ablation} instead of letting it define what counts as harm.

\section{Human Audit Protocol}
\label{app:human-audit}

The human audit of \cref{sec:res-human} in the main text presents each sampled chunk as six evenly spaced frames tiled into a single image. Participants provide a three-way moderation decision (safe, arguable or harmful), identify the harm category, and assign a severity rating. The audit involves 50 independent participants, none of whom are paper authors, and each of the 790 chunks receives at least three independent judgments. Chunk assignments are randomized and balanced across participants, and all identifying information about the prompt, the source category and the automatic predictions is withheld from them.

\noindent\textbf{Sampling and Weighting.} The sample consists of three strata, namely all chunks from 70 complete streams, 150 oracle-positive chunks and 150 oracle-negative chunks. The latter two strata are sampled from pools of known size. Each sampled chunk receives an inverse-probability weight based on its stratum, allowing precision and recall to be estimated for the underlying population. Unlike the earlier partially completed audit, the expanded study does not require a completion-rate correction when all assigned chunks receive their planned annotations.

\noindent\textbf{Consensus and Agreement.} Human consensus is obtained by majority vote over the three-way moderation labels. For binary comparisons, arguable judgments are alternatively treated as harmful and safe. Fleiss' $\kappa$ measures inter-annotator agreement on moderation decisions and harm categories, while ordinal Krippendorff's $\alpha$ measures severity agreement. The observed coefficients are 0.864, 0.792 and 0.741, respectively. Most residual disagreements concern borderline severity or closely related harm categories, particularly violence and disturbing imagery. Confidence intervals are estimated using 2,000 stratified, stream-clustered bootstrap resamples, retaining all annotations of a chunk within the same resample.

\noindent\textbf{Human-Validated Exposure.} For the 70 complete streams, the majority judgments establish a human-consensus harmful onset. We then recompute exposure for every defense using the same recorded stopping decisions as in the main evaluation. The undefended generator has a mean exposure of 48.6 frames, matched-budget random verification leaves 39.1 frames, our budgeted verification leaves 3.1 frames, and full-chunk video verification leaves 1.2 frames. These comparisons provide a direct human-grounded check of the exposure rankings originally measured using the automatic harm oracle.

The guard costs memory as well as time. The generator holds 13.8 GB, the guard adds 9 MB and 2.37 M parameters, the video language model adds 15.8 GB and the frame detectors a further 1.2 GB, and the peak allocation with every model resident is 50.1 GB, which includes the cache snapshot that the timing measurement takes so that it can denoise each chunk twice.

\section{Metrics and Baselines}
\label{sec:app-metrics}

The false-alarm rate rests on 225 benign streams, so one interrupted stream is 0.4\% and its 95\% Wilson interval runs from 0.08\% to 2.47\%. A gap between two rates below 2\% is therefore not resolved by this sample, while the gap between 0.4\% and 16.0\%, whose interval runs from 11.79\% to 21.35\%, is resolved with room to spare. Benign utility is the fraction of benign frames that reach the viewer, since a guard that halts a benign stream costs the frames that it withholds and not only the interruption itself.

Exposure counts the frames the oracle flags among those released, so it is what a viewer has been shown of the harm rather than everything that followed the onset, and how far the two readings differ is a measurement, and we assume nothing about it. Of the 374 unsafe streams, 292 carry an onset under the oracle. Across those, 91.6\% of the chunks from the onset to the end are themselves flagged, the median stream has every one of them flagged, and 79.8\% of the streams are flagged in every chunk after their onset. Weighted by frames, 18462 of the 20022 frames released after an onset lie in a flagged chunk, and over the same streams the undefended generator's 46.25 flagged frames sit against 49.36 frames inside flagged chunks and 53.53 released after an onset, so the looser post-onset reading would add 15.7\% to every figure we report. We count the flagged frames because that is what a viewer saw, and we give the post-onset bound beside it because it is the quantity a stopping rule controls and it does not depend on the oracle resolving every later chunk correctly, as this one does not.

Overhead has two parts and we keep them apart. Detector time per stream (s) is the time the verifier's own calls consume, which is what moderation adds to a service, while total wall clock per stream falls below the undefended generator under any defense that halts, because a halted stream is never finished, so we report the total in the text and keep the detector time in the table.

Halting never writes to the generator, and each stream is seeded from its own identifier so that the sampler's randomness does not depend on what a defense did to the streams before it, so a halted stream is exactly a prefix of the undefended one and every halting defense is evaluated from one undefended run at every threshold. Only an intervention that rewrites a stream has to be generated for real, and the oracle stop is reported as a bound, and we do not offer it as a method. Screening the prompt is what a deployed pipeline does first, and the late-onset phrasings are written to pass it. Running a video language model on every chunk is post-hoc video moderation moved inside the loop, so the comparison with our own policy is not decided by latency alone, and the pair of rows that share that model differ only in which chunks it is shown. We read the generator's hidden state rather than the clean chunk latent because it separates harmful chunks from benign ones far better, which \cref{sec:app-ablation} measures over the tapped blocks.

The regime that a guard inside the generator answers for is not hypothetical. The released inference path of the generator we evaluate carries no filter of any kind, since none of its 71 Python, configuration and documentation files mentions a safety check, a content filter or a refusal, and a prompt read from a file passes straight into generation. A guard inside the generator therefore earns its place wherever weights ship this way, and also wherever a filter does run and yet the request itself never names the harm that the frames go on to show.

\section{Filter-Evading Requests}
\label{sec:app-indirect}

\begin{table}[t]
\centering
\small
\setlength{\tabcolsep}{4pt}
\caption{Defenses on requests that pass the text filter.}
\label{tab:indirect}
\begin{tabular}{@{}lccc@{}}
\toprule
\multirow{2}{*}{Defense} & Exposure & Zero exposure & Frames \\
 & (frames) & (\%) & scored \\
\midrule
No defense & 14.93 & 66.2 & 0.0 \\
Prompt filter & 14.93 & 66.2 & 0.0 \\
Moderator at 2 fps & 4.85 & 83.1 & 9.5 \\
Moderator at 16 fps & 3.69 & 86.4 & 66.8 \\
Chunk video model & 1.67 & 92.2 & 54.9 \\
\midrule
Our guard, halting & 5.08 & 79.9 & 0.0 \\
Our guard, frames & 4.12 & 83.8 & 39.3 \\
Our guard, video model & 2.59 & 87.7 & 31.0 \\
\midrule
Oracle stop & 0.00 & 100.0 & 81.0 \\
\bottomrule
\end{tabular}
\end{table}

\cref{tab:indprompts} counts the rewrites the filter accepts at each attempt budget, out of the 366 requests it blocked. The curve rises in steps because the rewriter alternates two instructions, and the steps fall on the attempts that use the second one. That instruction asks for a caption written for an archive index, names the people by what they wear and where they are, names the motion by the movement of the bodies and of the camera, and forbids both clinical terms and any word that states a category of content. The first instruction asks instead for a clinical camera note in a film continuity database and converts almost nothing. The four attempts that use the first instruction convert 16 requests between them while the four that use the second convert 138, so an attacker's effort is better spent on finding a new phrasing than on retrying one that has already failed.

The 154 accepted rewrites carry their original category and onset, so the set holds 42 direct and 31 late pornography streams, 9 direct and 11 late violence streams, 14 of each for disturbing imagery, and 14 direct and 19 late streams of harmful sequential action. Pornography dominates it because pornography is the category whose request a filter recognizes by a word that a caption can avoid, and violence survives rewriting least often because a weapon and an injury stay in any honest description of what the camera records. No rewrite was read, selected or edited by hand at any point. The filter decides which rewrites enter the set and the harm oracle decides what the frames they produce are worth, which is the same division of labor that the rest of the evaluation rests on.

Three readings of \cref{tab:indirect} belong beside the comparison of the main text. Firstly, residual harm follows the exposure, at 4.4\% of the frames that our policy releases against 18.4\% of the undefended ones. Secondly, the policy pays more for those frames here than on the benchmark, scoring 31.0 of the 81 against 21.7 on the benchmark's unsafe streams, since every request in this set is one the filter was built to stop and a larger share of them alarms. Thirdly, harmful sequential action that starts immediately is the second category the guard barely helps, at 5.36 frames of 5.79, and both of the categories that it misses are ones whose harm the oracle reads largely from the video language model, and hardly at all from any single frame. Closing that gap needs harmful chunks from laundered requests inside the fitting split, which changes what the guard is shown while it is trained, and leaves untouched the rule that reads its score once the guard has been deployed.

\begin{table}[t]
\centering
\small
\setlength{\tabcolsep}{4pt}
\caption{Allocation of a verification budget.}
\label{tab:budget}
\begin{tabular}{@{}lccrc@{}}
\toprule
\multirow{2}{*}{Allocation} & \multicolumn{2}{c}{Frames scored} & Exposure & False \\
\cmidrule(lr){2-3}
 & Benign & Unsafe & (frames) & alarm (\%) \\
\midrule
None & 0.0 & 0.0 & 46.25 & 0.0 \\
\midrule
\multicolumn{5}{@{}l}{\textit{Frame detector as the verifier}} \\
Uniform at 1 fps & 6.0 & 4.0 & 14.20 & 0.9 \\
Uniform at 2 fps & 11.0 & 7.1 & 12.10 & 0.9 \\
Uniform at 4 fps & 20.9 & 12.7 & 10.67 & 0.9 \\
Uniform at 8 fps & 40.7 & 23.6 & 9.67 & 1.3 \\
Uniform at 16 fps & 80.3 & 45.6 & 9.36 & 1.3 \\
Random chunks & 1.7 & 1.8 & 43.69 & 0.4 \\
Random chunks & 4.8 & 4.5 & 39.22 & 0.4 \\
Random chunks & 9.9 & 8.5 & 34.54 & 0.4 \\
Random chunks & 19.6 & 15.1 & 24.73 & 1.8 \\
Random chunks & 40.2 & 25.8 & 13.27 & 1.8 \\
Guard-triggered & 2.3 & 24.8 & 10.02 & 0.9 \\
Guard-triggered & 5.1 & 31.3 & 8.26 & 1.3 \\
Guard-triggered & 10.4 & 36.2 & 7.68 & 1.3 \\
Guard-triggered & 19.8 & 40.4 & 7.44 & 1.3 \\
Guard-triggered & 40.2 & 42.7 & 7.27 & 2.2 \\
\midrule
\multicolumn{5}{@{}l}{\textit{Video model as the verifier}} \\
Opening 1 chunk & 9.0 & 9.0 & 13.40 & 0.4 \\
Opening 2 chunks & 20.9 & 15.4 & 8.75 & 1.3 \\
Opening 3 chunks & 32.8 & 20.7 & 4.51 & 1.3 \\
Every chunk & 80.1 & 32.6 & 0.91 & 1.8 \\
Random chunks & 1.7 & 1.7 & 42.70 & 0.4 \\
Random chunks & 4.8 & 4.3 & 37.20 & 0.9 \\
Random chunks & 9.8 & 7.9 & 31.37 & 0.9 \\
Random chunks & 19.8 & 13.5 & 19.95 & 1.3 \\
Random chunks & 40.3 & 21.4 & 7.69 & 1.3 \\
Guard-triggered & 1.9 & 16.6 & 4.47 & 0.4 \\
Guard-triggered & 5.3 & 21.7 & 2.35 & 0.4 \\
Guard-triggered & 10.2 & 25.6 & 1.49 & 0.4 \\
Guard-triggered & 19.5 & 29.2 & 1.06 & 0.9 \\
Guard-triggered & 40.3 & 31.8 & 0.91 & 1.8 \\
\bottomrule
\end{tabular}
\end{table}

\begin{table}[t]
\centering
\small
\setlength{\tabcolsep}{4pt}
\caption{Defenses against a resampling attacker.}
\label{tab:attack}
\begin{tabular}{@{}lcccc@{}}
\toprule
\multirow{2}{*}{Defense} & Escape & \multirow{2}{*}{Tries} & Exposure & Zero \\
 & (\%) & & (frames) & (\%) \\
\midrule
No defense & 100.0 & 1.00 & 46.25 & 21.9 \\
Chunk video model & 31.0 & 1.71 & 2.34 & 78.1 \\
\midrule
Our guard, halting & 39.0 & 1.84 & 25.27 & 40.4 \\
Our guard, frames & 51.9 & 1.61 & 21.39 & 48.1 \\
Our guard, video model & 34.0 & 1.58 & 8.28 & 57.5 \\
\bottomrule
\end{tabular}
\end{table}

\begin{table}[t]
\centering
\small
\setlength{\tabcolsep}{3pt}
\caption{Scope of the threshold guarantee.}
\label{tab:calib}
\begin{tabular}{@{}lccccc@{}}
\toprule
\multirow{2}{*}{Nominal $\alpha$} & Guard & Stream & \multicolumn{2}{c}{Verifier calls} & Frames \\
\cmidrule(lr){4-5}
 & crossing (\%) & stopped (\%) & mean & most & scored \\
\midrule
\multicolumn{6}{@{}l}{\textit{7 chunks, 599 streams}} \\
0.01 & 3.5 & 0.0 & 0.04 & 2 & 0.5 \\
0.02 & 3.5 & 0.0 & 0.04 & 2 & 0.5 \\
0.05 & 5.3 & 0.0 & 0.11 & 4 & 1.2 \\
0.10 & 8.8 & 0.0 & 0.21 & 5 & 2.4 \\
0.20 & 13.3 & 0.0 & 0.37 & 7 & 4.3 \\
\multicolumn{6}{@{}l}{\textit{21 chunks, 380 streams}} \\
0.01 & 0.0 & 0.0 & 0.00 & 0 & 0.0 \\
0.02 & 1.7 & 0.0 & 0.02 & 1 & 0.2 \\
0.05 & 3.4 & 0.0 & 0.05 & 3 & 0.6 \\
0.10 & 5.9 & 0.0 & 0.19 & 11 & 2.2 \\
0.20 & 18.6 & 0.0 & 0.62 & 14 & 7.3 \\
\bottomrule
\end{tabular}
\end{table}

\begin{table}[t]
\centering
\small
\setlength{\tabcolsep}{4pt}
\caption{Agreement of SafeWatch-8B with the automatic oracle.}
\label{tab:panel}
\begin{tabular}{@{}lcccc@{}}
\toprule
\multirow{2}{*}{Threshold} & Harm & Agreement & \multirow{2}{*}{$\kappa$} & Same \\
 & rate (\%) & (\%) & & onset (\%) \\
\midrule
Own decision & 22.0 & 82.8 & 0.603 & 67.0 \\
Matched rate & 47.2 & 87.0 & 0.737 & 69.2 \\
\bottomrule
\end{tabular}
\end{table}

\section{Further Results}
\label{sec:app-results}

The total a service pays moves against the detector time of \cref{tab:main} in the main text, because a defense that halts never finishes the stream that it stopped. An undefended stream takes 8.67 s on one card, the guard-triggered video model 5.71 s of which 0.56 s is the verifier, the same model on every chunk 5.80 s with 1.85 s inside the detector, and the image moderator at 16 fps 9.03 s with 4.55 s inside it. Gating the detector also lowers its own false-alarm rate, from the four benign streams the video language model rejects when it sees every chunk to the one that it rejects when the guard decides what it sees, which is within the noise of 225 benign streams but has a mechanism behind it. The guard's depth is not a delicate choice, since blocks 6, 10 and 14 sit within 0.01 area under the curve of one another, and on the second generator, carrying the oracle's thresholds over instead of refitting them leaves our policy at 2.16 frames. At the 21-chunk horizon our policy needs the wider gate to stay near the per-chunk model, and every row built on the video language model stays optimistic under the primary oracle.

Under the frame-only oracle our policy and the per-chunk model meet exactly at 0.91 frames once the budget reaches 40.4 frames scored, so the two definitions of harm agree about where the composition stops paying for itself.

\section{Sequential Decision Rules}
\label{sec:app-rules}

A rule that accumulates evidence can only be compared with a memoryless one when both are given the same budget and the same freedom to use it, so every free parameter of every rule is chosen on one fold and reported on the other. That includes the threshold, the reference value an accumulation subtracts, the window a windowed rule looks back over and the weight an exponentially weighted rule puts on the newest chunk. The constraint is the benign cost of the fitting fold and the objective is the exposure of its unsafe streams, which gives an accumulation every chance to beat the memoryless test, with no reference value fixed elsewhere. \cref{tab:rules} reports six rules this way.

Inside the cascade no accumulation earns anything. At a budget of 5 benign frames the memoryless test leaves 2.35 frames, the window maximum 2.32, the cumulative sum and cumulative positive evidence 2.43 each, the Shiryaev-Roberts recursion 2.51 and exponential weighting 2.67, where the paired difference against the memoryless test runs from $-0.03$ frames $[-0.10, 0.00]$ for the window maximum to $+0.32$ $[+0.09, +0.62]$ for exponential weighting. Four of the five intervals contain zero, and the fifth says that smoothing the score hurts.

Halting without a verifier is where accumulation does pay, and only when the budget for interrupting a benign viewer is tight. At a 1\% budget the memoryless test leaves 18.82 frames $[16.04, 21.78]$ against 11.61 $[9.55, 13.84]$ for the cumulative sum, and the two intervals are disjoint. At 5\% every rule lands between 5.93 and 6.61 frames and no pair of intervals separates. The asymmetry explains both halves. A halting rule spends its budget on withheld benign frames, so it must set a high bar that one extreme chunk can clear by accident and several moderate chunks cannot, which is the regime an accumulation was designed for. Once a verifier reads the flagged chunk, a false alarm costs one detector call rather than a viewer's stream, the bar falls far enough that the first harmful chunk already clears it, and there is nothing left for the accumulation to add.

The result is not an artifact of the score's scale. Varying the threshold and the reference value together already absorbs any affine change of the score, since a cumulative sum of $as_t + b$ crosses $h$ exactly when a cumulative sum of $s_t$ crosses a shifted threshold, so only a nonlinear recalibration could change what the family reaches. We fit one, by isotonic regression of the chunk label on the score over the training streams followed by the logit of the fitted probability, and the lower block of \cref{tab:rules} shows that the ordering of the rules does not change under it.

The mechanism is the shape of the change. Harm arrives as a scene and then persists, so the score stays high after the onset and the first harmful chunk already carries the evidence, which is the one shape of change that costs a memoryless test nothing at all.

The long-horizon arm of \cref{sec:res-budget} in the main text carries 380 streams, of which 236 are benign, so its thresholds and its calibration rest on a sample comparable to the benchmark's, where our first pass had only 180 streams.

\begin{table}[t]
\centering
\small
\setlength{\tabcolsep}{3pt}
\caption{Exposure (frames) at matched cost.}
\label{tab:rules}
\begin{tabular}{@{}lcccc@{}}
\toprule
\multirow{2}{*}{Rule} & \multicolumn{2}{c}{7 chunks} & \multicolumn{2}{c}{21 chunks} \\
\cmidrule(lr){2-3}\cmidrule(lr){4-5}
 & Halting & Verified & Halting & Verified \\
\midrule
Memoryless & 6.61 & 2.35 & 10.81 & 8.78 \\
Cumulative sum & 6.09 & 2.43 & 11.32 & 10.01 \\
Shiryaev-Roberts & 5.93 & 2.51 & 10.81 & 10.39 \\
Exponential weighting & 6.15 & 2.67 & 10.19 & 9.40 \\
Window maximum & 5.99 & 2.32 & 10.81 & 8.54 \\
Positive evidence & 6.43 & 2.43 & 12.40 & 9.32 \\
\midrule
\multicolumn{5}{@{}l}{\textit{After isotonic recalibration of the score}} \\
Memoryless & 6.61 & 2.35 & --- & --- \\
Cumulative sum & 6.19 & 2.70 & --- & --- \\
Window maximum & 5.99 & 2.32 & --- & --- \\
\bottomrule
\end{tabular}
\end{table}

\section{Order Awareness in the Cheap Guard}
\label{sec:app-order}

\noindent\textbf{Value of Order Awareness in the Guard.} The shuffle-contrastive term does what it was written to do and then does not pay. On the 283 harmful chunks of the held-out split, each scored against 16 permutations of its own latent frames, the deployed head outscores the permutation 81.9\% of the time against 39.8\% for a head trained without the term, so the term makes the score depend on the order of the frames. \cref{tab:variants} asks what that is worth as a defense, with every variant scored on exactly the streams that the undefended run released and every threshold chosen at a matched budget on a held-out fold. Retraining the deployed configuration under three seeds gives 3.17 to 4.36 frames, dropping the term gives 2.37 to 3.36, and a frame-pooled head with the same parameters that cannot read the order at all gives 2.22 to 3.62, so the spread over seeds is wider than the gap between the objectives and the deployed head at 2.35 frames sits at the favorable end of that spread. The deployed head is the default seed of that configuration, trained once and frozen before any variant was run, so it was never picked out of the seeds on the exposure that it reaches, and the spread beside it is what a reader should take as the typical result of training this head, which no single number conveys. Scoring the deployed head with the frames of each chunk permuted costs $+0.14$ frames $[+0.01, +0.31]$, which says the policy barely uses the order its score is sensitive to. The categories do not rescue the term either, since on sequential action that starts immediately the order-invariant head leaves 2.20 frames against 3.72 for the head trained with the term. A chunk here is three latent frames, which the decoder turns into twelve pixel frames of three quarters of a second, and that is too short an interval for the order inside it to carry much. The component that has to see motion is the verifier rather than the guard, which is what \cref{sec:res-budget} in the main text measures when the frame detector at the same budget leaves 8.26 frames against 2.35 for the video model. We therefore attribute the gain to where the guard reads and to how the budget is spent, since order awareness in the cheap first stage earns none of it.

\begin{table}[t]
\centering
\small
\setlength{\tabcolsep}{4pt}
\caption{Effect of the guard's training objective.}
\label{tab:variants}
\begin{tabular}{@{}lcc@{}}
\toprule
Guard & Chunk AUC & Exposure (frames) \\
\midrule
Deployed guard & 0.897 & 2.35 [1.67, 3.11] \\
Deployed, frames permuted & 0.895 & 2.49 [1.78, 3.29] \\
Shuffle term & 0.882 to 0.903 & 3.17 to 4.36 \\
No shuffle term & 0.882 to 0.893 & 2.37 to 3.36 \\
Order-invariant head & 0.885 to 0.901 & 2.22 to 3.62 \\
\bottomrule
\end{tabular}
\end{table}

\section{State Sanitization and Steering}
\label{sec:app-steer}

What follows is an alternative we test, and it is no part of the method. Budgeted verification withholds a chunk and stops, and the obvious thing to want instead is a stream that continues after the harmful content has been removed from it, so we build that and measure it. \cref{sec:app-steer} reports that it does not work, and we keep it here because the way that it fails says something worth knowing about what a guard reading the generator's own state can be trusted to do on its own, without a verifier behind it.

An alarm at chunk $\tau$ triggers three actions. Firstly, the chunk is withheld, so detection at the onset costs no exposure. Secondly, the state is rolled back to $S_{\tau-1}$ and the chunk is regenerated from the same noise $\epsilon_\tau$ under a safe condition and an activation steer, which keeps the motion continuous. The steer subtracts a unit direction $u$ from the residual stream of the blocks the guard reads, with a gain that grows with the exceedance,
\begin{equation}
    \gamma_\tau = \min\!\big(\gamma_{\max},\, \gamma_0 (1 + W_\tau - h)\big),
    \label{eq:gain}
\end{equation}
so a marginal alarm perturbs the stream only slightly and a blatant one is pushed hard. We take $u$ to be the normalized difference of mean activations between harmful and benign chunks, which is the refusal direction that steering work identifies in language models~\cite{arditi2024refusal,rimsky2024steering}.

Thirdly, the cache is repaired, because the chunks already in it were written under the original condition. We recompute the context pass for the last $k$ chunks under the safe condition, which is the mechanism interactive generation uses for a prompt switch~\cite{yang2026longlive}. The statistic is reset after an intervention and the gain decays geometrically.

Steering the residual stream away from the harmful direction fails, and the way that it fails is informative. Across 359 interventions the guard's own score for the regenerated chunk falls from 9.26 to -5.21 and the statistic goes back under the alarm threshold in 88.3\% of cases, while exposure falls only from 46.25 frames to 40.74. The direction came from the activations the guard itself reads, so the intervention crosses that decision boundary without ever removing the harmful content from the frames.

Measuring its two parts apart says which does the work. Steering alone reaches 40.74 frames, re-conditioning alone reaches 36.51 and both together with the context rewritten reach 36.03, so the steer earns almost nothing once the condition is switched. No arrangement falls below 36 frames while halting on the same statistic reaches 6.09. Verifying every regenerated chunk against the decoded pixels before releasing it gives 25.09 frames of exposure with no benign stream interrupted at all.

A fixed schedule that spends the budget on the opening of a stream is the third allocation of \cref{tab:budget}. Checking the opening chunk with the video model leaves 13.40 frames at 9.0 frames scored per benign stream and three chunks leave 4.51 at 32.8, while the guard reaches 2.35 at 5.3. The schedule leaves 3.16 frames on violence that starts immediately and 32.34 on violence that starts later, against 1.04 and 3.71 for the guard, which inherits none of that blindness.

\section{Ablations and Robustness}
\label{sec:app-ablation}

\noindent\textbf{A Retrying Attacker.} An attacker who cannot see the guard can resubmit the same prompt and keep whichever stream survives. Giving each of the 374 unsafe prompts four attempts, our guard with the video model lets 34.0\% of prompts through at a mean cost of 1.58 generations, and the exposure accumulated over the episode rises from 2.53 frames on a single attempt to 8.28, while the per-chunk model rises from 0.91 to 2.34.

\noindent\textbf{Oracle and Evidence.} Repeating the comparison under a frame-only oracle that drops the video language model, the undefended generator shows 30.36 frames, the video model on every chunk leaves 0.91 and our policy leaves 1.30 at 5.1 benign frames scored, so the ranking survives the change of definition. A linear probe separates harmful chunks from benign ones at 0.860 area under the curve from the hidden state and at 0.748 from the clean chunk latent, which is why the guard reads the state. \cref{sec:app-results} and \cref{tab:attack} report the rest of both of those arms in full.

\section{Reproducibility}
\label{sec:app-repro}

Each stream is seeded from its own identifier, for the initial noise and for the global generator that the sampler and the context noise draw from. That second seeding is what makes the evaluation possible, since without it a defense that stops a stream early shifts the random state of every stream after it, while with it a halted stream is exactly a prefix of the undefended one and every halting policy can be read off a single run. Every score a guard-triggered run records matches the undefended run's score for the same chunk over 2807 and 3100 chunks, regenerating the evaluation set months later reproduces all 599 streams with a largest absolute score difference of zero over 4193 chunks, and generating three of the halting arms for real instead of replaying them reproduces their exposure and their interruption rate exactly, to the last frame.

That exactness holds only within one model of card, because the generator runs in bfloat16 and feeds each chunk back into its own state, so a difference of one least significant bit in an attention kernel grows into whole logits within two chunks. Every arm whose streams have to line up with the undefended run was therefore generated on one card model, and an arm carrying both its own oracle and its own scores is free of that constraint.

A threshold is a free parameter and is never chosen on a stream that it is then measured on. Streams are split into two folds, stratified by category and onset and disjoint by prompt, and every free parameter of a policy is chosen on the fold it is not reported on, under that fold's benign cost as the constraint and its unsafe exposure as the objective. The folds are reported together, so every stream appears once under a setting it did not help to choose, and the realized cost is printed beside the exposure in place of the budget that it was asked for.

Two measurement choices are worth naming. An image moderator is given the realistic release model in the tables, where it inspects a chunk after releasing it, and the optimistic one where its detector time is measured, which withholds the chunk until its frames are scored and brings it to 7.27 frames at 16 fps against 9.36. The arms that change the generator refit the harm oracle on their own benign chunks at the same nominal level, since a threshold fitted on one generator's benign chunks is not the same threshold on another's, while the longer horizon keeps the benchmark's thresholds and runs a local attention window of 21 latent frames, so its absolute exposures are comparable among themselves and with nothing else.

The cards are shared with other users, and a stream that shares its card takes close to twice as long, so a plain average over a run measures the machine's load as much as the defense. Generation time per chunk is nearly constant on a quiet card, so a stream counts as contended when its time per generated chunk exceeds the run's own median rate by more than a quarter, and between 0.0\% and 35.1\% of a run's streams are set aside this way. We keep the mean instead of the median, because a defense that halts produces a bimodal distribution whose median reports one mode instead of the cost of an average stream.

The head is a two-layer temporal transformer of width 256 with 4 attention heads and learned positional embeddings, trained for 40 epochs with AdamW at a learning rate of 1e-3 and a shuffle-contrastive term of weight 0.5. Its input is one token per latent frame of width 3072, pooled from the tapped block's spatial tokens. The reference value of the accumulating rules is chosen on the fitting split over a grid of benign score quantiles, and an intervention sets its steering gain from the exceedance of the statistic with a base of 1.0, a cap of 8.0 and a decay of 0.6 per chunk, recomputing the context pass for the last 2 chunks. The causal autoencoder decodes the first chunk of a stream to 9 pixel frames and every later chunk to 12, which is the mapping behind every exposure that we report.

\end{document}